%% file: main.tex
\documentclass[10pt,twocolumn,letterpaper]{article}

\usepackage{times}
\usepackage{epsfig}
\usepackage{graphicx}
\usepackage{amsmath}
\usepackage{amssymb}
\usepackage{caption}
\usepackage{pifont}
\usepackage{arydshln}

\usepackage{subcaption}
\usepackage{tabularx}
\usepackage{makecell}
\usepackage{multirow}

\usepackage{mathtools}
\usepackage{booktabs}
\usepackage{xcolor,colortbl}
\usepackage[accsupp]{axessibility}
\definecolor{mygray}{gray}{0.9}
\definecolor{myblue}{rgb}{0.54, 0.81, 0.94}
\definecolor{myblue2}{rgb}{0.67, 0.9, 0.93}

\input{macros}

\usepackage{wacv}              

\usepackage{graphicx}
\usepackage{amsmath}
\usepackage{amssymb}
\usepackage{booktabs}

\usepackage[pagebackref,breaklinks,colorlinks]{hyperref}
\usepackage[capitalize]{cleveref}

\newcommand\blfootnote[1]{%
  \begingroup
  \renewcommand\thefootnote{}\footnote{#1}%
  \addtocounter{footnote}{-1}%
  \endgroup
}

\crefname{section}{Sec.}{Secs.}
\Crefname{section}{Section}{Sections}
\Crefname{table}{Table}{Tables}
\crefname{table}{Tab.}{Tabs.}

\def\wacvPaperID{2435} 
\def\confName{WACV}
\def\confYear{2024}
\begin{document}

\title{CrashCar101: Procedural Generation for Damage Assessment}


\author{
Jens Parslov$^{*,1}$ \quad \quad 
Erik Riise$^{*,1}$ \quad \quad 
Dim~P.~Papadopoulos$^{1,2}$ \\
$^1$ Technical University of Denmark \quad \quad 
$^2$ Pioneer Center for AI \\
{\tt\small jens@parslov.com, erikriise@live.no, dimp@dtu.dk}  \\
\url{https://crashcar.compute.dtu.dk}
}
\maketitle
\input{0_Abstract}
\input{1_Introduction}
\input{2_Related_Work}

\input{3_Synthetic_Data_Generation}\input{4_Model}
\input{5_Experiments}
\input{6_Conclusion}

\newpage

{\small
\bibliographystyle{ieee_fullname}
\bibliography{egbib}
}

\end{document}

%% file: macros.tex
\iftrue     
\newcommand{\dimpp}[1]{\textcolor{red}{[DP: #1]}}
\newcommand{\jens}[1]{\textcolor{magenta}{[JP: #1]}}
\newcommand{\erik}[1]{\textcolor{cyan}{[ER: #1]}}
\else
\newcommand{\dimpp}[1]{\textcolor{red}{\noindent}}
\newcommand{\jens}[1]{\textcolor{magenta}{\noindent}}
\newcommand{\erik}[1]{\textcolor{orange}{\noindent}}
\fi

\newcommand{\mypar}[1]{\vspace{-0mm}\noindent\textbf{#1}}

\newcommand{\xmark}{\ding{55}}
\newcommand{\cmark}{\ding{51}}%

%% file: 0_Abstract.tex
\begin{abstract}

In this paper, we are interested in addressing the problem of damage assessment for vehicles, such as cars. This task requires not only detecting the location and the extent of the damage but also identifying the damaged part. To train a computer vision system for the semantic part and damage segmentation in images, we need to manually annotate images with costly pixel annotations for both part categories and damage types. To overcome this need, we propose to use synthetic data to train these models. Synthetic data can provide samples with high variability, pixel-accurate annotations, and arbitrarily large training sets without any human intervention. We propose a procedural generation pipeline that damages 3D car models and we obtain synthetic 2D images of damaged cars paired with pixel-accurate annotations for part and damage categories. To validate our idea, we execute our pipeline and render our CrashCar101 dataset. We run experiments on three real datasets for the tasks of part and damage segmentation. For part segmentation, we show that the segmentation models trained on a combination of real data and our synthetic data outperform all models trained only on real data. For damage segmentation, we show the sim2real transfer ability of CrashCar101.
\end{abstract}

%% file: 1_Introduction.tex
\section{Introduction}
\blfootnote{$^*$Denotes equal contribution}%
Damage assessment is the task of determining the extent of damage resulting from an accident or a disaster. Vehicle damage is a common type that occurs from transportation and road accidents~\cite{patil2017, weber2020detecting, weber2022incidents1m}. Damage assessment is essential for response organizations, insurance businesses, and rental agents. This is usually performed by experts who manually check vehicles on-site and evaluate their damages.

Recently, there have been several attempts to build automatic damage assessment systems with computer vision models~\cite{patil2017,zhang2020automatic,weber2020detecting,cardd2022,weber2022incidents1m}. 
However, automatically assessing damages on vehicles is a challenging task. It requires not only detecting and localizing the specific damages on the vehicles but also accessing the extent of the damage depending on the part of the vehicle which is damaged. 
From a computer vision perspective, this means going beyond the image-level classification problem and training a semantic segmentation model able to produce per-pixel category predictions for both damage types and semantic parts in new test images of vehicles. 
Training such a model requires huge amounts of annotated images where humans draw detailed outlines around every damage and part that appears in an image~\cite{cordts16cvpr,gupta19cvpr,lin14eccv,russell08ijcv,zhou17cvpr}. This process is expensive and prone to several erroneous labels especially when the annotation task is challenging~\cite{northcutt2021confident,northcutt2021pervasive,sambasivan2021everyone,shankar2020evaluating}.

\begin{figure}[t]
\center
\includegraphics[width=1\linewidth]{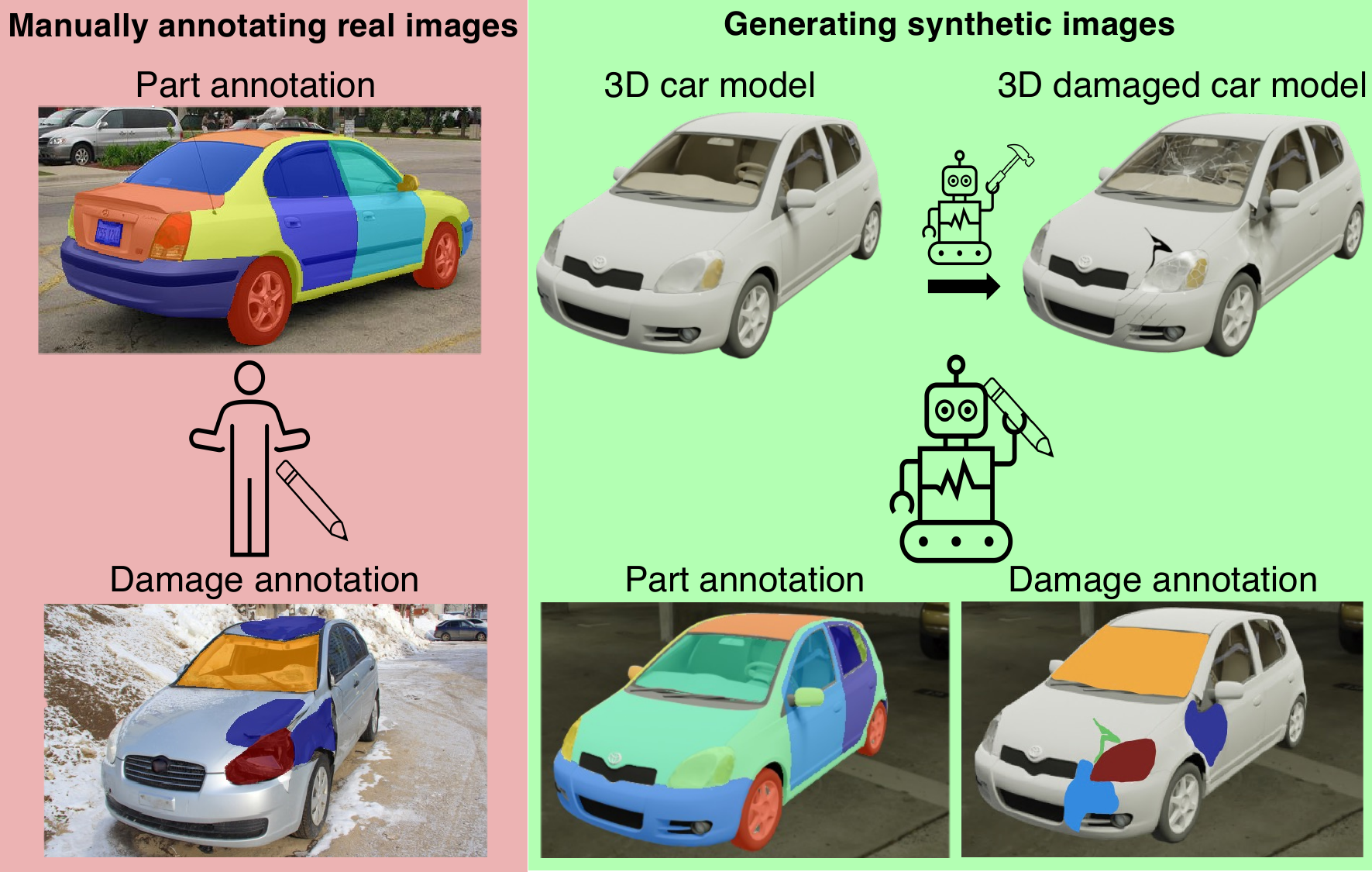}
\caption{ \small {\textbf{Learning part and damage segmentation from synthetic data.} (Left) The standard approach of manually annotating real car images with pixel annotations. (Right) We propose to use synthetic 3D car models, destroy them using a procedural damage generation pipeline, and obtain realistic 2D images that come with pixel-accurate annotations for semantic parts and damages.}}
\label{fig:teaser}
\end{figure}

\begin{figure*}[t]%
    \centering
    \includegraphics[width=1\linewidth]{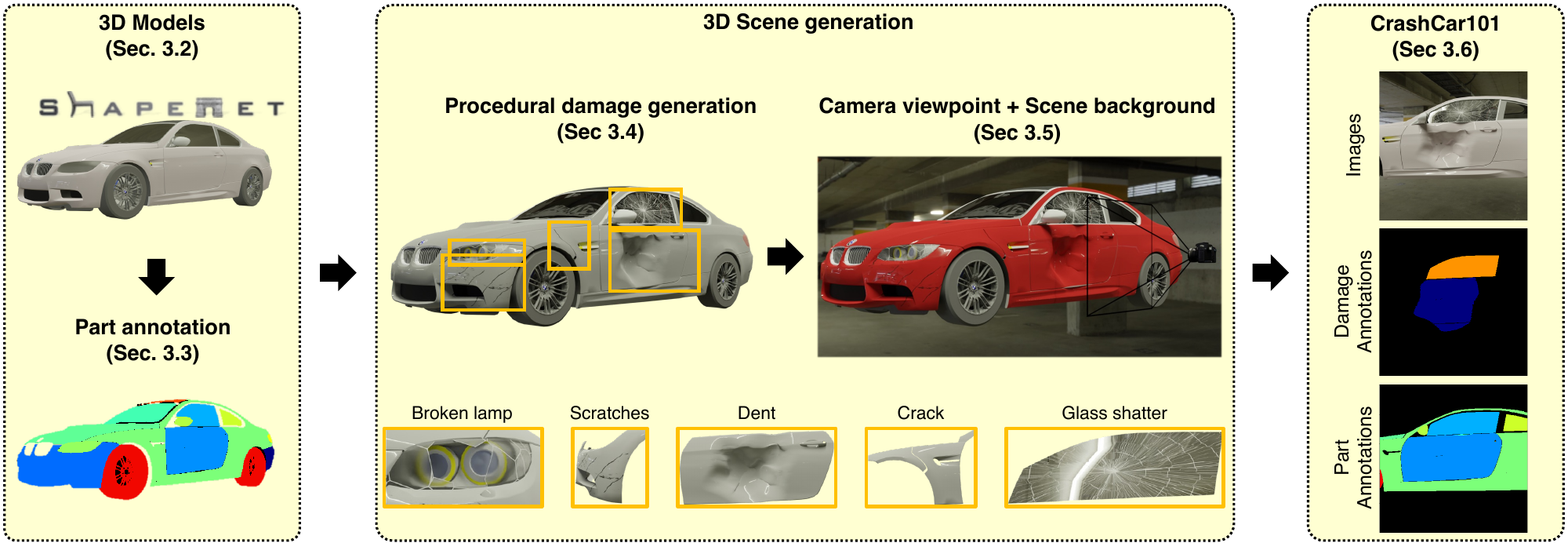}
    \caption{\small \textbf{Overview of our proposed procedural generation pipeline. } We first acquire and annotate 3D model cars (Sec.~\ref{sec:3_2} and~\ref{sec:3_3}), and then we apply procedural generation to manipulate the texture and shape of the car to generate synthetic damage types (Sec.~\ref{sec:3_4}. Subsequently, we place a camera into the scene and assign a scene environment and car color (Sec.~\ref{sec:3_5}). Finally, we render a 2D image paired with perfect ground-truth pixel annotations for parts and damages. Given a set of 3D model cars with part annotations, this process is fully automatic, allowing us to render an arbitrarily large amount of data (CrashCar101 dataset, Sec.~\ref{sec:3_6}).}
    \label{fig:syngen}
\end{figure*}

In this paper, we propose to overcome these issues by automatically generating realistic synthetic data to train a damage assessment system (Fig.~\ref{fig:teaser}). Realistic synthetic images have been used successfully in several semantic segmentation tasks~\cite{gaidon2016virtual,liu2022learning,procsy,ros16cvpr,sankaranarayanan2018learning}. Synthetic data can provide samples with high variability and perfect automatically generated pixel annotations. When they are generated via a procedural generation pipeline, they can lead to arbitrarily large sets of training data without any human intervention~\cite{de2017procedural,gaidon2016virtual,hewitt2023procedural,khan2019procsy}.
We propose a novel procedural damage generation pipeline that creates realistic damages of various types on 3D car models~\cite{shapenet2015} (Sec.~\ref{sec:synthetic_generation}). We focus only on cars as the most common vehicle type, but a similar procedure can be followed for other objects. We create damages by manipulating either the 3D mesh geometry (\textit{dents}) or the physically-based material of the models (\textit{scratches, cracks, shattered glass, broken lamps}). By controlling the model parameters, we can create damages at different scales, positions, shapes, and appearance variations. 

Our procedural generation pipeline is shown in Fig.~\ref{fig:syngen}. We start with a 3D car model~\cite{shapenet2015} where we annotate sub-meshes with semantic part labels. We apply our damage generation pipeline to obtain damages into the 3D car model. Then, we place the generated model into an urban HDRI scene that provides realistic lighting and background noise. Finally, we set the camera parameters in order to obtain a 2D image. The final image is paired with two generated segmentation maps with pixel-accurate labels, one for the car semantic parts and one for the damage types. 

To validate our idea, we first annotate 99 car models from ShapeNetCore~\cite{shapenet2015} with part annotations and then we execute our generation pipeline to obtain our CrashCar101 dataset that consists of 101,050 images paired with perfect part and damage segmentation. 
We conduct experiments on three real test datasets to show the usefulness of our synthetic dataset.
For damage segmentation, our results in a few-shot learning scenario show that pre-training on CrashCar101 yields significantly better results (+6.3-17.9\% mIoU at 1-shot and +4.4-7.0\% at 5-shot) compared to using a pre-trained model on COCO~\cite{lin14eccv} or ImageNet~\cite{deng09cvpr}.
For part segmentation, we show that the segmentation models trained on a combination of real data and our synthetic data outperform all models trained only on real data.

%% file: 2_Related_Work.tex
\section{Related work}

\mypar{Synthetic data} has been a valuable asset for addressing several problems such as semantic segmentation for urban landscapes~\cite{ros16cvpr, richter2016, Biasetton_2019_CVPR_Workshops}, object recognition~\cite{hattori2015learning,peng2015learning,Tremblay2018}, face-related tasks~\cite{wood2021fake} and medical imaging~\cite{FRIDADAR2018}. 
Procedural generation is an important field as it gives the ability to generate millions of 3D scenes without any human intervention~\cite{synthbook}. Procedural generation is widespread in video games~\cite{procgengames,bookprocgames,gengames}, but it has also been used for 
synthetic data generation~\cite{procsy,pose,graph}. For example, in~\cite{procsy} cities and outdoor scenes are generated for semantic segmentation.
In~\cite{graph}, a stochastic scene grammar from an indoor dataset is learned to generate new scene layouts. In~\cite{pose}, entire humans are generated and deep networks are fitted on the resulting images to regress a dense set of annotations. 



\begin{figure}[t]%
    \centering
    \includegraphics[width=\linewidth]{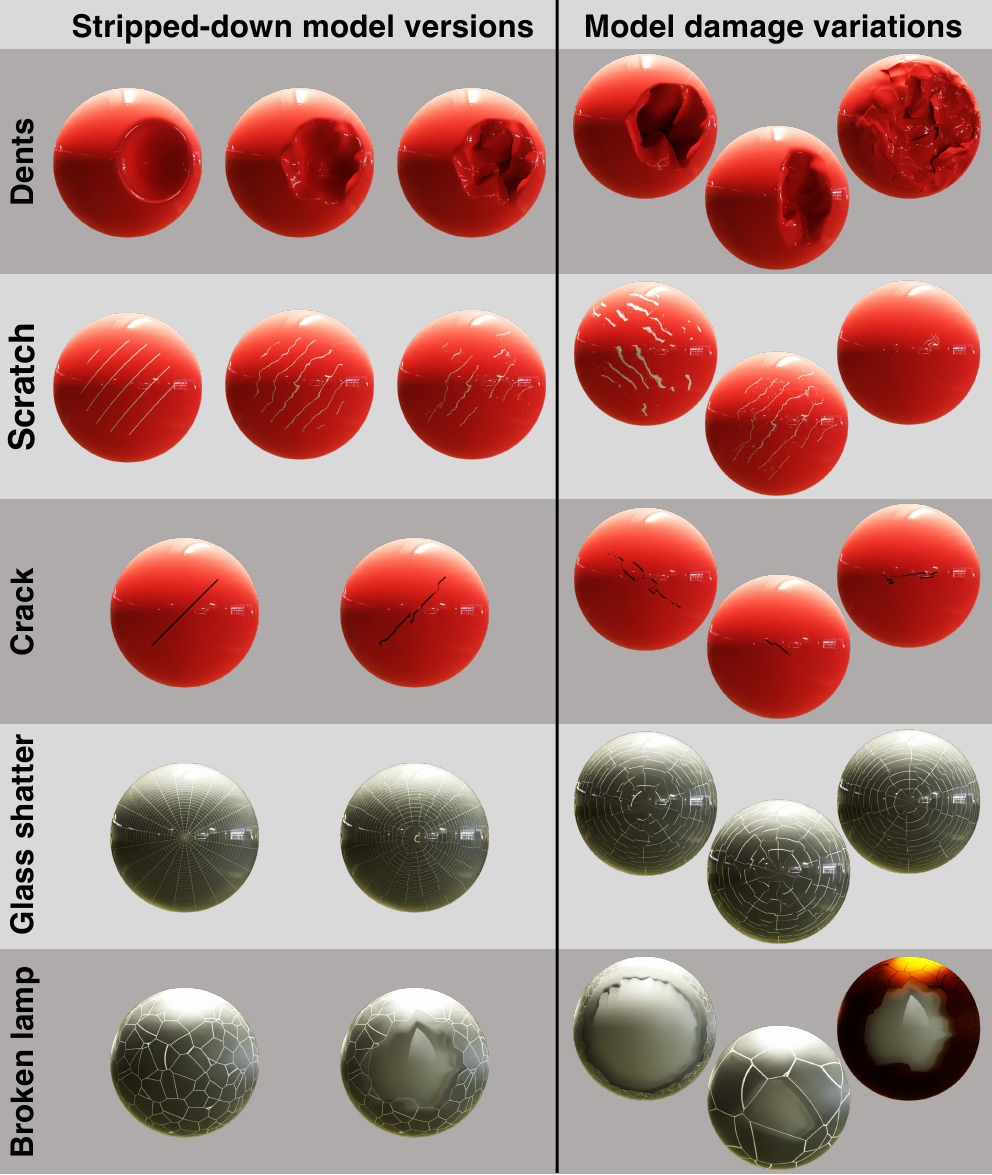}
    \caption{\small \textbf{Damage models on toy 3D spheres}. We show stripped-down versions of each model (left) and examples of how the different values of the model parameters affect the appearance and texture of the damage (right). The texture of the spheres resembles the texture of the corresponding car parts (dents, scratches and cracks on the body parts, glass shatters on the windows, and broken lamp on headlights.}
    \label{fig:bumps}%
\end{figure}

\mypar{Semantic part segmentation} is the task of segmenting fine-grained parts within a target object such as cars~\cite{liu2022learning, chen14cvpr, pasupa2021}, birds~\cite{Saha2022,wah2011}, humans~\cite{chen14cvpr} or buildings~\cite{wang2020}. Obtaining annotations for semantic parts is expensive and challenging since there is often no clear boundary to separate the object parts. For these reasons, the research has been focused on ways to reduce this manual annotation cost by training weakly supervised models~\cite{zhou2018}, unsupervised models~\cite{hung2019} or domain adaptation approaches on synthetic data~\cite{liu2022learning}.

No large-scale dataset for part segmentation on cars exists. Pascal-Part~\cite{chen14cvpr} includes 613 images with 13 part categories. CGPART~\cite{liu2019semantic} has 31 fine-grained part categories but only 40 images. In Sec.~\ref{sec:part_experiments}, we conduct experiments in both datasets and we show performance improvements for the task of part segmentation when using our synthetic data.

\mypar{Damage detection}  is an essential problem for emergency response especially in cases of natural disasters, destructive events, or accidents~\cite{gupta2019xbd,maeda2018,patil2017,weber2020detecting, weber2022incidents1m}.
Several papers have been published more closely related to our task, as they delve into the task of classifying damage in cars~\cite{vanRuitenbeek2022, mahavir2021,patil2017, cardd2022,Singh2019AutomatingCI, vehits19, patel2020, Li2018, dhieb2019, zhang2019}. However, with the exception of~\cite{cardd2022}, they all focus on image classification tasks.
To the best of our knowledge, there are only three publicly available datasets with labeled car damage \cite{cardd2022, neokt2017, LPLENKA2020}. \cite{neokt2017} contains only image-level labels, while ~\cite{LPLENKA2020} contains only 80 images where all damage types are annotated as one category. The CarDD dataset~\cite{cardd2022}, which was recently released, contains 4,000 images with pixel annotations for six damage types. In Sec.~\ref{sec:damage_experiments}, we run experiments in CarDD showing the usefulness of CrashCar101 for the task of damage segmentation on real images.

\mypar{Shape 3D models.}
The ShapeNet dataset~\cite{shapenet2015} is a large collection of 3D object models. ShapeNetCore~\cite{savva2016shrec16} is a subset of ShapeNet with models of 55 manually verified object categories, including cars. ShapeNetCore has been used in several computer vision applications such as aligning CAD models~\cite{Avetisyan2019},
predicting object-level intrinsics~\cite{shi2017} or generating shapes from natural language~\cite{chen2019text2shape}.
In~\cite{mo2019partnet}, PartNet extends a subset of ShapeNetCore~\cite{shapenet2015} with fine-grained 3D part segmentation over 24 object categories. However, none of these categories include vehicles.
In this paper, we annotate and provide fine-grained part segmentation for 27 semantic parts on 99 car models from ShapeNetCore~\cite{shapenet2015} (Sec.~\ref{sec:3_3}).


%% file: 3_Synthetic_Data_Generation.tex
\section{Procedural damage generation}
\label{sec:synthetic_generation}

In this section, we explain each step of our procedural synthetic data generation pipeline (Fig. \ref{fig:syngen}). All rendering, texturing and 3D mesh manipulation is conducted in the open-source software application Blender~\footnote{https://www.blender.org/}.

\subsection{Overview}
\label{sec:3_1}
Fig. \ref{fig:syngen} presents an overview of our procedural generation pipeline.
We first acquire a diverse set of 3D car models from the ShapeNetCore dataset~\cite{shapenet2015} (Sec.~\ref{sec:3_2}) and we manually label the sub-meshes of these models with fine-grained part categories (Sec.~\ref{sec:3_3}).
Then, we apply our proposed method for procedural damage generation which is presented in Sec.~\ref{sec:3_4}. After the damages are placed in the 3D car, we set the scene environment, and the car color and we place the camera into the scene leading to 2D images paired with pixel-accurate annotations for part categories and damage types (Sec.~\ref{sec:3_5}).
We execute this pipeline and we obtain and render the CrashCar101 dataset which consists of 101,050 images (Sec.~\ref{sec:3_6}).


\subsection{3D models}
\label{sec:3_2}

We use 3D vehicle models from ShapeNetCore~\cite{shapenet2015}. 
Even though the PartNet extends a subset of ShapeNetCore~\cite{shapenet2015} with 3D part segmentation over 24 object categories, none of these categories include vehicles.
As a result, in this paper, we manually annotate the sub-meshes of 99 selected 3D car models from ShapeNetCore with fine-grained part categories. We choose to use the part taxonomy from~\cite{liu2022learning} which consists of 31 part categories. We further combine all four wheel categories into one and the two license plate categories into one. This results in 27 part categories. 
%


\subsection{Part annotation}
\label{sec:3_3}

Manually annotating every sub-mesh of every car model and mapping it to a part category is expensive. 
To make this process more efficient, we follow a human-in-the-loop approach. We start by manually annotating nine models from ShapeNetCore. Then,  we obtain eight images for each annotated model from different viewpoints around the car and we use them to train a DeepLabv3 segmentation model~\cite{liang2017,He_2016_CVPR}.
For each new car model from ShapeNetCore, we render the same eight viewpoints and predict the semantic part segmentation using the trained model. 
The meshes of the 3D cars are compared with the predictions using mIoU computed across the 8 images. For each part class, we rank the meshes using mIoU from least to most likely. In order to utilize the ranking, we created an interactive labeling tool and captured eight views of each car. By employing the mIoU metric, the tool displays the eight most likely meshes belonging to each part. A single car model in ShapeNetCore can have thousands of sub-meshes~\cite{shapenet2015}, making manual labeling expensive and tedious.
According to our time recordings, this human-in-the-loop interactive approach was about 3$\times$ faster than the fully manual one.






\subsection{Procedural Generation for Synthetic Damage} 
\label{sec:3_4}

The goal of our procedural generation pipeline is to create realistic damages on 3D car models. 
We consider 5 damage types: \emph{dents}, \emph{scratches}, \emph{cracks}, \emph{broken lamps}, and \emph{glass shatters}, which reflect the most common damage types~\cite{cardd2022}.
 In this section, we present damage generators that manipulate the shape and texture of the car. These generators apply procedural rules that model damages in generality.  Each damage generator has interpretable input parameters that control the damage and can be sampled from pre-defined distributions to generate random variations of damages and lead to a final dataset with high variability.

\begin{figure*}[t]%
    \centering
    \includegraphics[width=\linewidth]{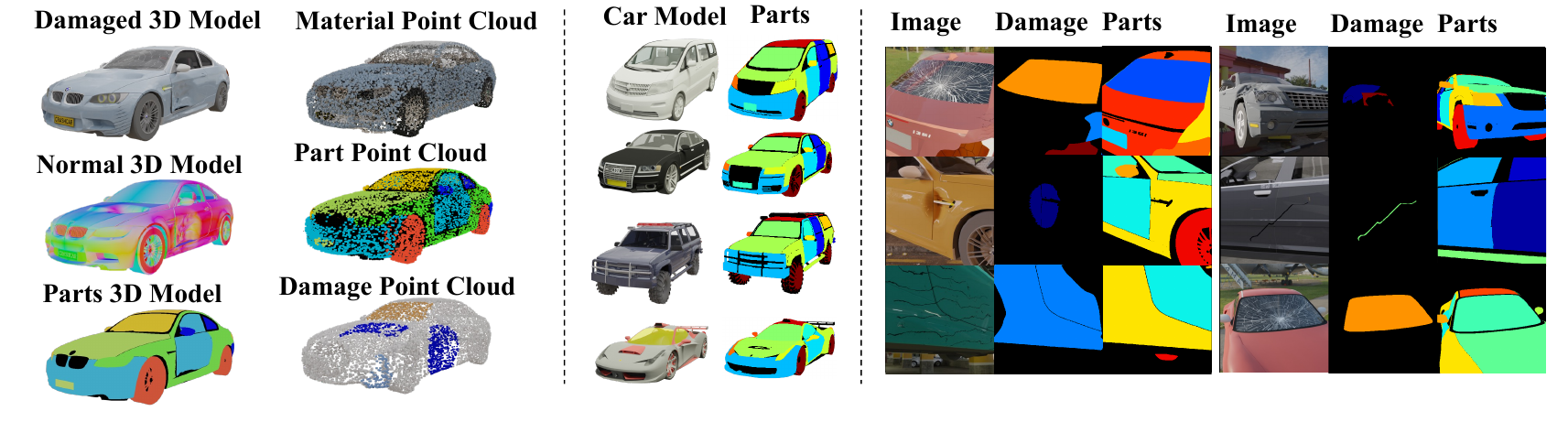}
    \caption{ \small \textbf{CrashCar101 Synthetic Dataset.} (Left) Depicts a 3D damaged car model, including its normals and part labels, from which we extract point cloud representations of material, parts, and damage. (Middle) Displays a subset of car models with labeled parts, while (right) showcases diverse 2D images generated from the 3D models.}
    \label{fig:dataset_examples}%
\end{figure*}

To better understand the damage generators, we apply them on toy 3D spheres in Fig.~\ref{fig:bumps}. We show stripped-down versions of each damage model (Fig.~\ref{fig:bumps} (left)) and how different input parameters affect the appearance and texture of the damage (Fig.~\ref{fig:bumps} (right)).
In the following $\epsilon_{w}$, $\epsilon_{v_{c}}$, $\epsilon_{v_{d}}$ and $\epsilon_{p}$ refer to Wave, Voronoi color, Voronoi distance and Perlin noise implementations of noise generators in 
Blender. 

\mypar{Dents} are generated using a function $f_D(x): \mathbb{R}^3\rightarrow \mathbb{R}^3$ that maps the vertex coordinates of an undamaged car to perturbed coordinates that depict a damaged car. The dent generator is implemented using Blender Geometry Nodes. The function $f_D(\textbf{x})$ consists of two components: $l(\textbf{x})$ and $d(\textbf{x})$. $l(\textbf{x})\in \mathbb{R}$ defines the length of the displacement vector, and $d(\textbf{x})\in \mathbb{R}^{3}$ defines the direction. The direction component $d(\textbf{x})$ is defined as $\textbf{n}+{\epsilon_{v_c}}(\textbf{x})$, where $\textbf{n}$ is the normal vector and $\epsilon_{v_c}: \mathbb{R}^3\rightarrow\mathbb{R}^3$ is a noise generator that simulates a crumpling effect by adding noise to the displacement direction. The effect is visualized in the second and third spheres in Fig.~\ref{fig:bumps} (first row). The length component $l(\textbf{x})$ is defined as $d \frac{\cos\left(\pi\frac{a-||\textbf{x}-\textbf{c}+\epsilon_p(\textbf{x})||}{a}\right)+1}{2}$, where \textbf{c} is the center coordinate of the dent, $a$ is the area of the dent, $d$ is the depth and $\epsilon_p:\mathbb{R}^3\rightarrow\mathbb{R}$ is a noise generator which adds noise to the magnitude of displacement as illustrated on the first and second sphere in Fig.~\ref{fig:bumps} (first row). The cosine function produces a gradual smooth transition into the dent.
A car paint shader \(S_p\) is applied to texture the car's body, while a glass shader \(S_g\) is used for the windows. The integration of the remaining damage types is achieved through Blender's Mix Shader node, blending two input shaders based on a probability factor. To facilitate this blending process, functions \(f:\mathbb{R}^3 \rightarrow [0,1]\) are defined, mapping coordinates on the car's surface to a probability. This probability determines the mixture of shaders, combining the undamaged car shaders \(S_p\) and \(S_g\) with shaders containing damage effects. Let $m(p,S_1, S_2)$ denote the Mix shader in Blender where shader $S_1$ and $S_2$ are being mixed with probability $p$.

\mypar{Scratches} are generated by overlaying $S_p$ with a scratch shader $S_d$ using generated scratch marks. To generate scratch intensities we define
\begin{align*}
f_r(\textbf{x}) &=
                   \left[ a \geq \|\mathbf{x} - \mathbf{c} + \epsilon_{v_{c}}(\mathbf{x})\| \right]
 \\
S_{scratch} & = m(f_r(\textbf{x})\epsilon_{w}(\textbf{x}),S_d,S_p)
\end{align*}

\noindent where [P] denotes the Iverson bracket notation and it equals 1 if the statement P is true and 0 otherwise. $f_r(\textbf{x})$ is used to select the scratched region of the car,  \textbf{c} determines the center, $a$ determines the area, and $\epsilon_{v_{c}}(\textbf{x})\in\mathbb{R}^3$ adds noise as seen on the first and third sphere in Fig.~\ref{fig:bumps} (second row). $\epsilon_w$ produces straight thin lines, which are manipulated using a distortion property as depicted on the first and second sphere in Fig.~\ref{fig:bumps} (second row).

\mypar{Cracks} are generated by mixing $S_p$ with a transparent shader $S_{\alpha}$ using generated crack masks. The crack mask is a line with a combination of smooth and sharp deviations. Let $x_1$ and $x_2$ be the first and second coordinates of $\textbf{x}$. We define $\hat{x}_{1}=x_{1}+\epsilon_{v_{c_{1}}}$,  $\hat{x}_{2}=x_{2}+\epsilon_{v_{c_{2}}}$ as noisy coordinates which produce deviation as shown on the first two spheres of Fig.~\ref{fig:bumps} (third row). We propose $f_{l}(\textbf{x})=1-\max(1-|\hat{x}_{1}-\hat{x}_{2}|,0)$  to produce a line and $f_r(\textbf{x})=\max(0,a-||\textbf{c}-\textbf{x}||)$ to select the crack region. 
\begin{align*}
S_{crack} &= m(f_{l}(\textbf{x})f_{r}(\textbf{x}), S_{\alpha}, S_p)
\end{align*}
\mypar{Glass shatter}
are generated by combining two shaders: a glass shader $S_{g}$ and a white non-transparent one $S_{w}$. Glass often shatters with lines radiating from the place of incident, alongside concentric spread rings. Once again $\epsilon_{v_{c}}$ is used to produce sharp deviations (see the first two spheres of Fig.~\ref{fig:bumps}(fourth row)).
To produce concentric rings we use  $f_{c}(\textbf{x})=\sin(||s(\textbf{x}-\textbf{c}+\epsilon_{v_{c}}(\textbf{x}))||)<t$ where $t$ defines the thickness, and $s$ defines the scale. For the radial lines, we use the radial gradient texture in Blender $\epsilon_{g}$ and use $f_{r}(\textbf{x})=\epsilon_{v_{d}}\left(\epsilon_{g}(\textbf{x})\right)||\textbf{x}||<t$, here $\epsilon_d$ is a noisy periodic function, which we use to determine the number of generated lines and adding noise to the distances between the lines. The shatter shader is defined as
$$S_{shatter}=m([f_{c}(\textbf{x})\textit{ or }f_{r}(\textbf{x})],S_{w},S_{g})$$

\mypar{Broken lamps} are generated by making a fractured shader $S_{f}$. To make $S_{f}$ we mix a glass shader $S_{g}$ with a white shader $S_{w}$. $S_{f}=m([\epsilon_{v_{d}}(\textbf{x})<t],S_{w},S_{g})$ where $t$ determines the thickness of the fractures, an example of $S_{f}$ can be seen at the first sphere of the last row in Fig.~\ref{fig:bumps}. A chunk is removed by mixing in a transparent shader $S_{\alpha}$ as depicted in Fig.~\ref{fig:bumps} (second sphere, last row), to produce the final broken lamp shader $S_{broken}$.
$$S_{broken} = m([||\epsilon_{v_{p}}(\textbf{x})||
<a], S_{\alpha}, S_f)$$

\mypar{Position of damage center}. To select the center of the damage $\textbf{c}$, we follow the following procedure: We first select randomly one main damage among the 5 damage types. Then, one of the car parts that can contain this damage is selected with equal probability, and a random vertex with coordinates $\textbf{c}_{main}$ from the part is selected. We set the parameter $\textbf{c}=\textbf{c}_{main}$ for the main selected type. All the other parameters are sampled from a uniform distribution. The minimum and maximum values are manually selected to be the most extreme values seen in realistic cases. 

\subsection{Camera viewpoint and scene background}
\label{sec:3_5}

The following section describes how the 3D scene is randomized to generate realistic synthetic 2D images.
First, we describe how the camera is placed, such that the damage is visible and then, we describe how the car paint and background are randomized to generate realistic 2D images.

\mypar{Viewpoint randomisation} is done by randomising the camera position $\textbf{v}$ and the camera rotation $\boldsymbol{\theta}$. Having placed the damage at some point $\textbf{c}_{main}$ we now place the camera position at $\textbf{c}_{main}$ and then translating in the direction of $R\left(\hat{n}_{\text {yaw }}, \theta_{yaw}\right) R\left(\hat{n}_{\text {pitch}}, \theta_{\text {pitch}}\right) \frac{v}{\|v\|}$ by some distance $d$. Where $R\left(\hat{n}_{\text {}}, \theta_{}\right)$ is the rotation matrix when $\hat{n}$ is the rotation axis and $\theta$ is the rotation angle. Note that we wish to control the pitch and yaw of the vector independently thus we select $\hat{n}_{\text{yaw}}=(0,0,1)$ and $\hat{n}_{\text{pitch}}=(-c_{main_2},c_{main_1},0)$. Finally we obtain the camera coordinates $\textbf{v}$ as:
\begin{equation*}   \textbf{v}=\textbf{c}_{main}+R\left(\hat{n}_{\text {yaw }}, \theta_{yaw}\right) R\left(\hat{n}_{\text {pitch}}, \theta_{\text {pitch}}\right) \frac{\textbf{c}_{main}}{\|\textbf{c}_{main}\|} \cdot d
\end{equation*}
Now that the camera is placed, we select $\boldsymbol{\theta}$ such that $\textbf{c}_{main}$ is perfectly centered in the frame. We now randomly select $\theta_1$ and $\theta_2$ in such a way that the main damage is jittered with respect to the raster coordinates.
Now that the primary damage is determined we apply secondary damage. To ensure that the secondary damage is visible all vertices of the parts that can contain damage are transformed to raster coordinates, and only those that are contained in the frame are kept. Only the vertices within a certain distance to the camera are kept. Finally, a random vertex of one of the visible objects is selected. This second damage is applied with a probability of 0.5, thereafter 0.2 until no damage is applied.


\mypar{Background randomization} After completing the annotation step, the scenes for which the vehicles are to be placed are initialized. We collected a total of 338 urban scene 4K HDRIs from Polyhaven~\footnote{https://polyhaven.com/hdris/urban}. The HDRI provides realistic lighting and background noise. Further, to add variation, we sampled realistic vehicle colors from GTA V~\footnote{https://wiki.rage.mp/index.php?title=Vehicle\_Colors\&oldid=21033}.

\subsection{CrashCar101 dataset}
\label{sec:3_6}

We execute our procedural generation pipeline and we render the CrashCar101 dataset. 
CrashCar101 consists of 101,050 2D images paired with annotated damage and part segmentation. Fig.~\ref{fig:dataset_examples} shows examples of CrashCar101. Our procedural generators damage the car, from which we can extract 3D and 2D modalities. From this, we produce CrashCar101, a 2D image dataset containing both part and damage segmentations. A subset of our dataset does not contain any damage, we refer to this subset as CrashCar101-Part. It consists of 17,325 images (175 images per car model) and we use it to train the part segmentation models in Sec.~\ref{sec:part_experiments}. We focus on 2D images, but generating 3D modalities is a feature available in the synthetic data generation pipeline. This is intended to enable further studies into the applicability of synthetic data in 3D research.
\begin{table}[t]
    \centering
    \resizebox{\linewidth}{!}{
    \begin{tabular}{l c c c c c }
         \bottomrule
        \rowcolor{myblue2}
        \textbf{Dataset}  & \textbf{Train} & \textbf{Val} & \textbf{Test} & \textbf{Part} & \textbf{Dmg} \\ \toprule
        Pascal-Part~\cite{chen14cvpr}  & 490 & 61 & 62 & \checkmark &  \\
        UDAPART~\cite{liu2022learning}  & - & - & 40 & \checkmark &  \\
        CGPART~\cite{liu2019semantic}  & 31,448 & 7,867 & - & \checkmark & \\ 
        \textbf{CrashCar101-Part}  & \textbf{14,175} & \textbf{1,575} &\textbf{1,575} & \checkmark &  \\
        \midrule
        CarDD~\cite{cardd2022} & 2,638 & 768 & 349 & & \checkmark \\
        \textbf{CrashCar101} & \textbf{83,604} & \textbf{8,311} & \textbf{9,135} & \checkmark & \checkmark \\ 
        \bottomrule
        
    \end{tabular}
    }
    \caption{\small \textbf{Datasets used in our experiments.} The top part of the table shows the datasets for the part segmentation experiments while the bottom part shows the ones for damage segmentation. The last two columns ``Part'' and ``Dmg'' indicate the existence of part and damage annotations in the dataset. }
    \label{tab:data}
\end{table}

%
Regarding the damage categories, we show interesting statistics of our obtained dataset (Fig.~\ref{fig:damage_stats}). In Fig.~\ref{fig:damage_size}, we show the distribution of each damage size in terms of the percentage of pixels they occupy in each image. As expected, we observe that cracks are usually tiny, while glass shatter damages usually occupy much larger image parts. 
The distribution of damage occurrence on each part is presented in Fig.~\ref{fig:dm_part_distribution} showing a good balance between damage types. In Fig.~\ref{fig:damage_distribution}, we show the number of images containing each separate damage in CrashCar101. Note that damages can only occur on specific parts (e.g., dents, cracks, and scratches on metallic parts, glass shatters on window glasses, and broken lamps on head and tail lights).

%% file: 5_Experiments.tex
\section{Experimental results}
\label{sec:experiments}
\begin{table}[t]
\centering 
\resizebox{\linewidth}{!}{
\begin{tabular}{c|lcccc}
\bottomrule
\rowcolor{myblue2}
 & \textbf{Dataset} & \textbf{Augs} &
 \multicolumn{2}{c}{\textbf{UDAPART}} &
 \multicolumn{1}{c}{\textbf{PASCAL-Part}} \\
 \rowcolor{myblue2}
 & \textbf{Parts} &  & \textbf{11} & \textbf{27} & \textbf{11} \\
\toprule
\multirow{8}{*}{\shortstack{DeepLabv3\\(RN50)}} & \multirow{2}{*}{Pascal-Part~\cite{chen14cvpr}} & \xmark & 42.1 & - &36.8 \\

& & \cmark & 37.1 & - & 36.0 \\ \cdashline{2-6}

& \multirow{2}{*}{CGPART~\cite{liu2019semantic}} & \xmark & 17.5 & 15.8 & 4.8 \\

& & \cmark & 34.3 & 25.7& 14.4 \\ \cdashline{2-6}

& \multirow{2}{*}{CrashCar101-Part} & \xmark & 39.3 & \textbf{42.8}  & 19.2 \\

& & \cmark & 43.0 & 41.5  & 22.4 \\ \cdashline{2-6}

& CrashCar101-Part + & \xmark & 43.1 & - & 36.0 \\
 
 & Pascal-Part~\cite{chen14cvpr} & \cmark & \textbf{52.5} & - & \textbf{42.9} \\

\bottomrule

\multirow{8}{*}{\shortstack{DeepLabv3\\(RN101)}} & \multirow{2}{*}{Pascal-Part~\cite{chen14cvpr}} & \xmark & 40.9 & - & 36.8 \\

& & \cmark & 40.0 & - & 37.3 \\ \cdashline{2-6}

& \multirow{2}{*}{CGPART~\cite{liu2019semantic}} & \xmark & 17.8 & 15.6 & 5.6 \\

& & \cmark & 38.1 & 30.1& 13.5 \\ \cdashline{2-6}

& \multirow{2}{*}{CrashCar101-Part} & \xmark & 40.4 & \textbf{47.6} & 20.0  \\

& & \cmark & 50.1 & 47.2  & 26.4 \\ \cdashline{2-6}

& CrashCar101-Part + & \xmark & 45.3 & - & 39.2 \\
 
 & Pascal-Part~\cite{chen14cvpr}  & \cmark & \textbf{52.3} & - & \textbf{44.2} \\

 \bottomrule
 
 \multirow{8}{*}{\shortstack{Segformer\\(b5)}} & \multirow{2}{*}{Pascal-Part~\cite{chen14cvpr}} & \xmark & 40.0 & - & 37.1 \\
 
& & \cmark & 38.0 & - & 36.7 \\ \cdashline{2-6}

& \multirow{2}{*}{CGPART~\cite{liu2019semantic}} & \xmark & 27.6 & 19.2& 8.0 \\

& & \cmark & 52.4 & 52.7& 24.8 \\ \cdashline{2-6}

& \multirow{2}{*}{CrashCar101-Part} & \xmark & 46.4 & 48.8  & 26.0 \\

& & \cmark & \textbf{56.3} & \textbf{61.6}  & 31.1 \\ \cdashline{2-6}

& CrashCar101-Part + & \xmark & 45.1 & - & 41.3 \\
 
 & Pascal-Part~\cite{chen14cvpr} & \cmark & 55.2 & - & \textbf{45.6} \\

 \bottomrule
\end{tabular}
}
\caption{\small \textbf{Part segmentation mIoU results.} We report the mIoU performance for each experiment with and without augmentations. For each test set (column) and model, we highlight in bold the best performance.}
\label{tab:partresults}
\end{table}


This section presents our experimental results. We evaluate the usefulness of our CrashCar101 dataset on two tasks: semantic part segmentation (Sec.~\ref{sec:part_experiments}) and damage segmentation (Sec.~\ref{sec:damage_experiments}). For each task, we compare the segmentation models trained on our dataset to models trained on real images and on combinations of real and synthetic data.

\mypar{Implementation details.} Unless stated otherwise, we use the following settings for all models of our two tasks.
All models are trained with the focal loss function~\cite{focalloss}. The focal loss is a modification of the standard cross-entropy loss that assigns higher weights to hard-to-classify examples, leading to improved performance on imbalanced datasets.
For damage segmentation, we use 6 output dimensions (5 damage types plus the background category).
For the part segmentation, we use either 28 or 12 output dimensions (including background) depending on which real dataset we evaluate the part segmentation models (see Sec.~\ref{sec:part_experiments} for more details). 
%
The input images are resized to 256$\times$256 for the task of part segmentation and 384$\times$384 for the task of damage segmentation.
We use a batch size of 64 for all models.
For a fair comparison among all trained models, we perform the same set of augmentations while training. These are limited to random resize cropping, random rotation, and color jitter. For part segmentation models, we also train models without augmentations to evaluate the effect of augmentations on our data compared to real data.
We train all models for 20 epochs using the Adam optimizer~\cite{kingma15iclr} with an initial learning rate of 0.0002. The only exception is the model trained on Pascal-Part, which due to the small size of the dataset was trained for 40 epochs.
We perform early stopping based on the performance in the hold-out validation set. Note that each different training set has its corresponding validation set from the same domain.
All experiments were run on a single Nvidia V100 GPU. 

\begin{figure}[t]\centering
\subfloat[The distribution of damage area by damage type.]{\label{fig:damage_size}\includegraphics[width=.49\linewidth]{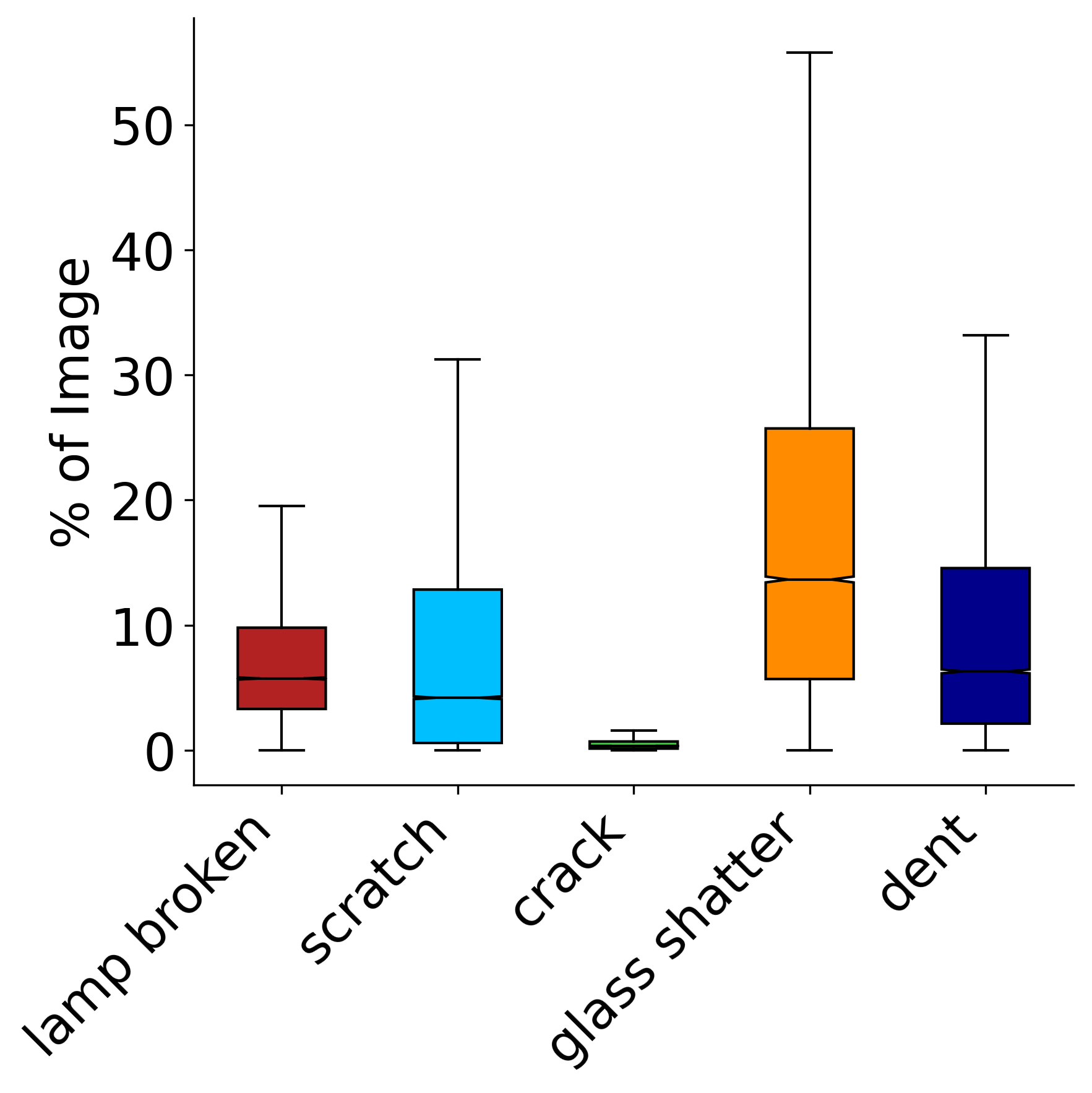}}\hfill
\subfloat[The number of images containing each damage.]{\label{fig:damage_distribution}\includegraphics[width=.49\linewidth]{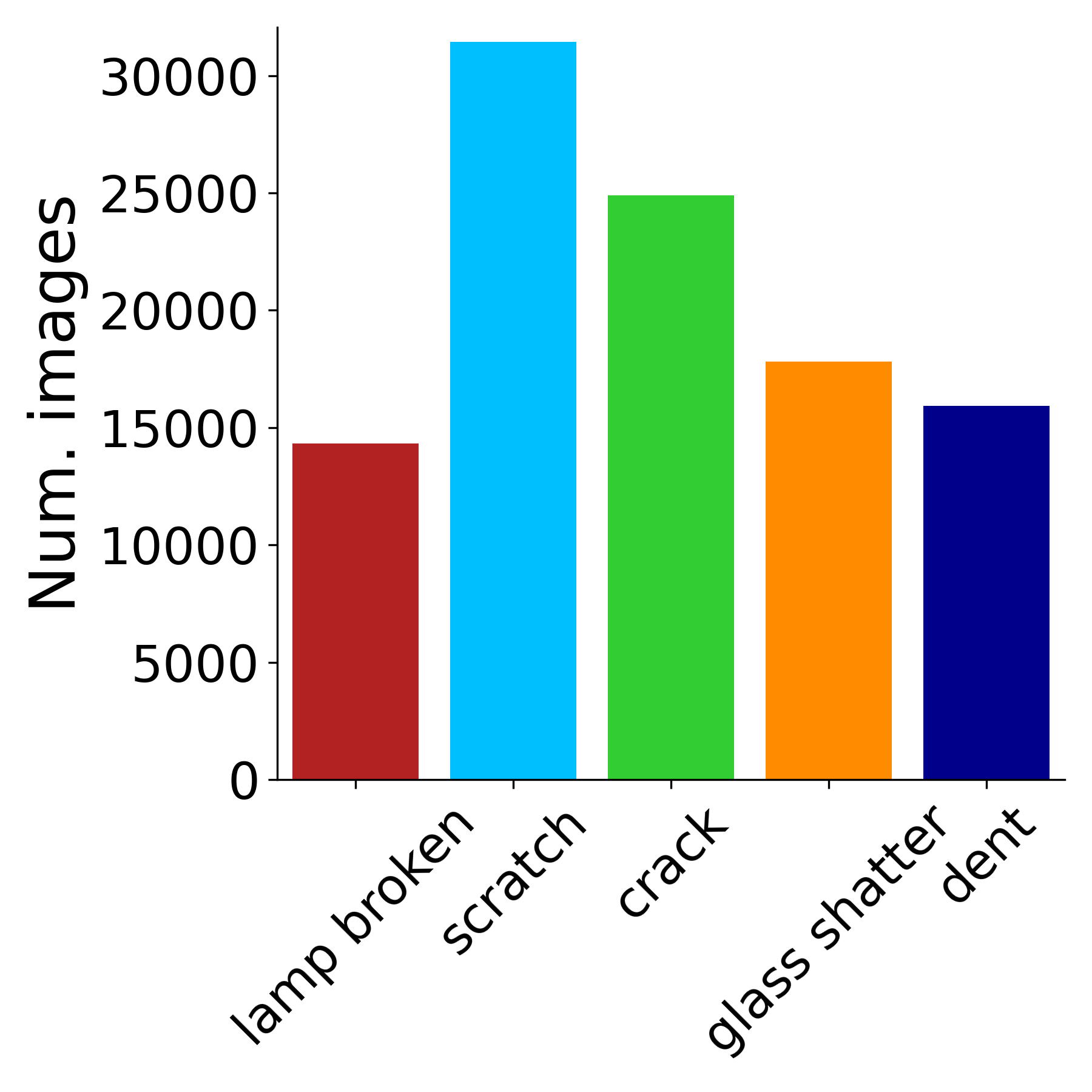}}\par 
\subfloat[Distribution of damage types appearing on car parts.]{\label{fig:dm_part_distribution}\includegraphics[width=1\linewidth]{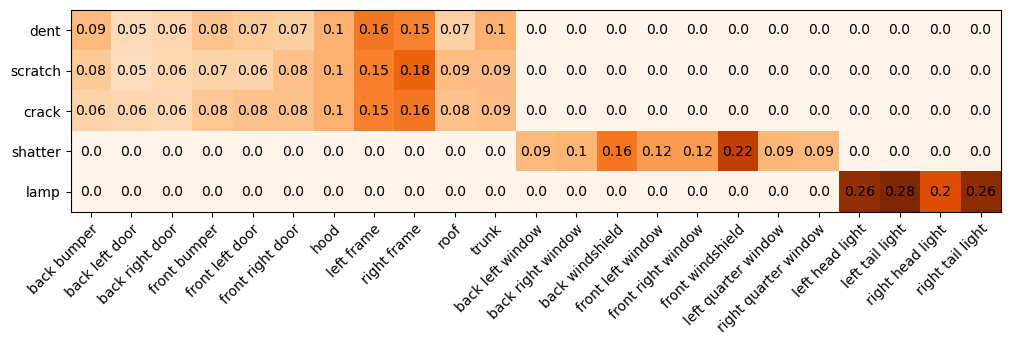}}
\caption{ \small \textbf{Damage statisics in CrachCar101.} 
(a) The distribution of damage area for each damage. We observe the minimal area of cracks compared to the more extensive area of shattered glass. (b) The number of images containing every damage. It demonstrates a uniform initial damage selection, highlighting a subsequent preferential selection of more suitable damage types on visible parts. (c) The distribution of damage types on car parts. We observe a size-dependent occurrence of damage on distinct parts, wherein larger components exhibit heightened damage incidence.}
\label{fig:damage_stats}
\end{figure}

\mypar{Evaluation.}
We use the mean intersection-over-union (mIoU) as our main evaluation metric for both tasks, as this is standard for evaluating any image segmentation task.


\subsection{Part segmentation}
\label{sec:part_experiments}

\mypar{Datasets.}
We use three publicly-available datasets: Pascal-Part~\cite{chen14cvpr}, UDAPART \cite{liu2022learning} and CG-PART \cite{liu2019semantic}. An overview of these datasets can be seen in Tab.~\ref{tab:data}.
Pascal-Part contains part annotations for 15 object categories on the Pascal VOC 2010 images~\cite{everingham10ijcv}. We crop cars around their bounding boxes and keep those with at least 16,384 pixels and less than 75\% background. This results in 612 images. The \textit{license plate} category was removed due to the poor annotation and the limited representation. The pixels corresponding to this category are labeled as \textit{front} or \textit{back} category depending on their location. 
%
For UDAPART and CG-PART, we merge the four wheel classes to one and the two license plates to one to align with our part definition. 

\mypar{Evaluation sets.}
For evaluating our models, we use two test sets with real images: Pascal-Part~\cite{chen14cvpr} and  UDAPART~\cite{liu2022learning}.
 The Pascal-Part original test set consists of 62 test images and we evaluate our models using the annotations with 11 semantic parts. UDAPART consists of 40 images and we use the whole dataset for evaluating our models using the annotations with 27 semantic parts. To enable the training and testing across these datasets, we also evaluate models on UDAPART by merging the 27 fine-grained categories to the 11 coarser categories of Pascal-Part. 

 \begin{figure}[t]%
    \centering
    \includegraphics[width=\linewidth]{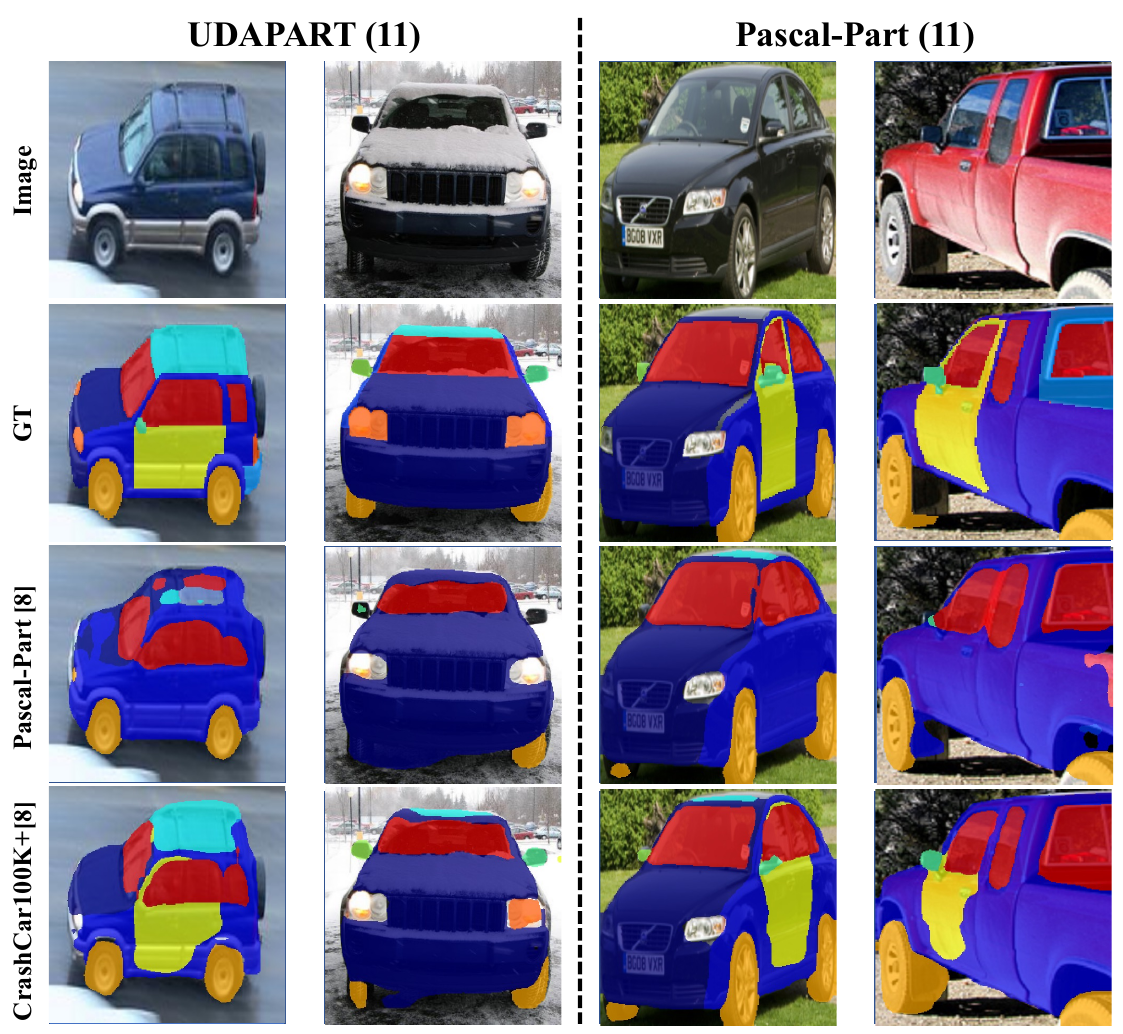}
    \caption{ \small \textbf{Qualitative part segmentation results.} We show results from training on Pascal-Part and on  CrashCar101+Pascal-Part using DeepLabv3 with a ResNet101 backbone. Both models were trained with augmentations (lines 10 and 16 in Tab~\ref{tab:partresults}). We observe that by including our synthetic data to the real training set, we obtain a  model that yields better results.}
    \label{fig:examples_parts}%
\end{figure}

\mypar{Training sets.}
We use four training sets to train our models:
(a) the \emph{Pascal-Part} training set with 490 real images, 
(b) the \emph{CGPART}~\cite{liu2019semantic} training set with 31,448 synthetic images, 
(c) our \emph{CrashCar101-Part} training set with 14,175 synthetic images, and
(d) the combination of the \emph{CrashCar101-Part $+$ Pascal-Part} training sets.
We train 18 part segmentation models in total using these sets. We train 12 models, three for each training set, with 11 part categories. We also train six more models with 27 part categories when the Pascal-Part set is not used (i.e., using the sets (b) and (c)).

\mypar{Segmentation models.}
We employ three semantic segmentation models with and without augmentations. We utilize DeepLabv3~\cite{liang2017} with ResNet50 and ResNet101 backbones~\cite{He_2016_CVPR}, both pre-trained on ImageNet~\cite{deng09cvpr}. Additionally, we use a B5-sized SegFormer model~\cite{xie2021} pretrained on Cityscapes~\cite{cordts16cvpr}. This approach provides a comprehensive evaluation of our dataset's part segmentation potential across diverse architectures and pre-training sources.


\mypar{Results.}
We report the results of the 18 trained models with augmentations and without on our evaluation sets in Tab.~\ref{tab:partresults}. As expected, we observe that the augmentations make a substantial difference in the results of synthetic data (especially in the case of \textit{CGPART} \cite{liu2019semantic}). 
We observe that our synthetic data outperforms \textit{CGPART}, the other state-of-the-art synthetic dataset in every instance. Even though the \textit{CGPART} training set is about double in size compared to our training set (see Tab.~\ref{tab:data}), we show that our dataset is much better and more realistic due to our rendering procedure and the number of the different car models used (99 models in CrashCar101-Part vs. 6 models in CGPart).

Moreover, we observe that the most precise models are those trained on the combination of real and synthetic data. Still, when utilizing the Segformer model, our CrashCar101-Part dataset outperforms even the model trained on the combination.

Interestingly, we observe that when evaluating on the real images of UDAPart, the model trained only on our synthetic data (third row for each respective model) significantly outperforms the one trained on real images of Pascal-Part (top row of each respective model). In the case of Deeplabv3 using the ResNet101 backbone and Segformer models, there's a noticeable performance boost (+9.2-16.3\% mIoU). We find the effect of the augmentations to be particularly interesting here. The synthetic data is not able to outperform real data without augmentations for the DeepLabv3 models, but when the training images are augmented, it outperforms even the real data trained without augmentations. Meanwhile, using the Segformer synthetic data performs better than real data outright. In Fig~\ref{fig:examples_parts}, we show qualitative test examples on both real datasets.

\subsection{Damage segmentation}
\label{sec:damage_experiments}

\mypar{Damage dataset and evaluation set.}
We use the recently released CarDD dataset~\cite{cardd2022} which contains real images of damaged cars annotated with object segmentation masks for several damage categories. To align with the damage types of our synthetic data, we remove the category flat tires. Images containing only flat tires are filtered out and the pixels annotated as flat tires are set to background. An overview of the dataset can be seen at the bottom part of Tab.~\ref{tab:data}. We evaluate our models in this section on the CarDD test set~\cite{cardd2022} which consists of 349 manually annotated images. 

\mypar{Experimental setup.} We evaluate the CrashCar101's sim2real transfer potential on damage segmentation using few-shot segmentation (FSS). FSS aims to segment novel objects with few annotations. Recent approaches~\cite{Yangetal,Zhangetal,Shenetal,tangetal} mitigate limited data by freezing the backbone, leveraging feature fusion and prototypes. For simplicity’s sake, we perform FSS experiments in a similar fashion to the baseline method in~\cite{chen2019}. We start by training on a large source dataset, namely COCO~\cite{lin14eccv}, ImageNet~\cite{deng09cvpr}, or CrashCar101. We then fine-tune the model on $n\cdot k$ images from the CarDD training set and we use an internal validation set consisting of $n\cdot k$ to perform early stoppage. The parameter $n$ denotes the number of shots and $k=6$ is the number of classes (including background). The $n\cdot k$ images are selected such that all class labels are present. We experiment with various freezing strategies and we report results obtained without freezing because these perform best. To quantify the reduction in domain gap, we compare the pre-training on CrashCar101 to the one on COCO and ImageNet respectively.

\mypar{Damage segmentation models.}
To show that the efficacy of CarCrash101 is architecture-independent, we perform FSS experiments using SegFormer and DeepLabV3 frameworks, each utilizing MiT-b5 and ResNet-101  backbones respectively. We adapt these models for damage segmentation by modifying the final layers to yield 6 channels, corresponding to distinct damage types.


\begin{figure}[t]
    \centering
    \includegraphics[width=.49\linewidth]{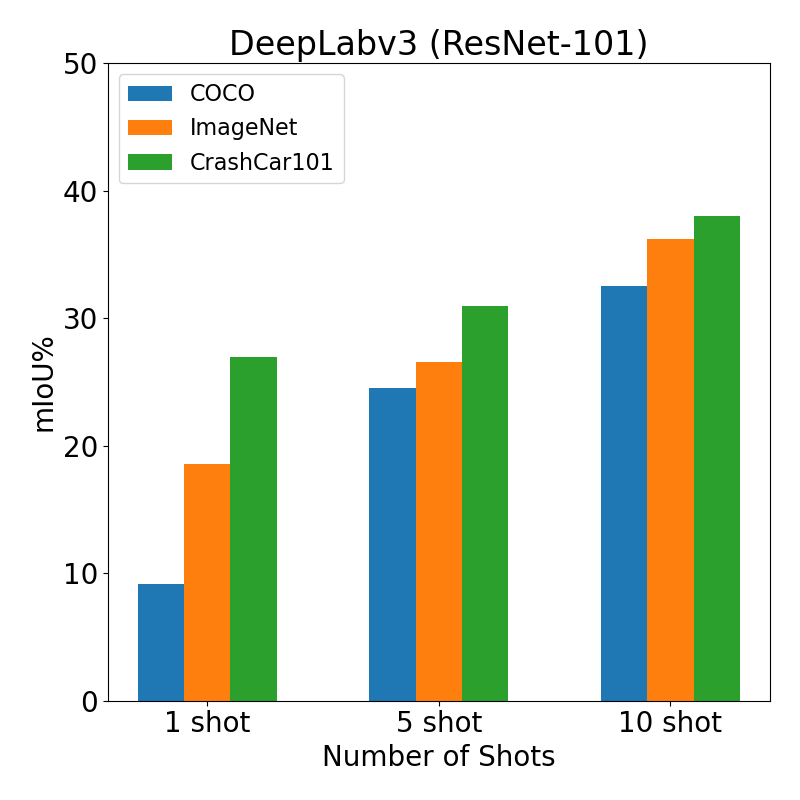}\hfill
    \includegraphics[width=.49\linewidth]{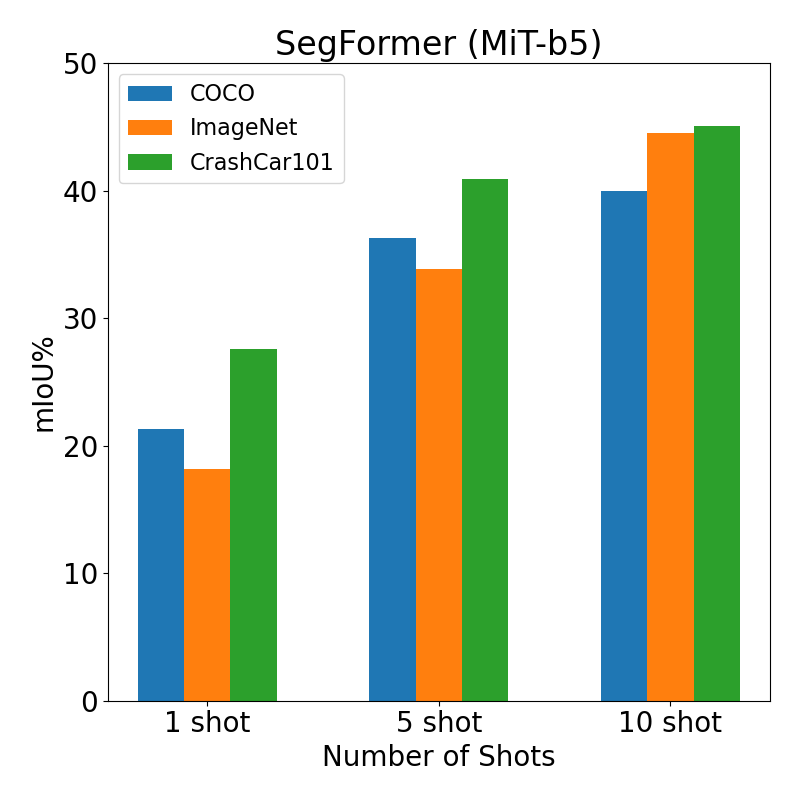}\par
    \caption{\textbf{Few-shot segmentation on damage segmentation using different model architecture}. Left: DeepLabv3 and right: SegFormer. Models pre-trained on CrashCar101 consistently outperform others regardless of the model architecture and without using any domain adaptation techniques.}
    \label{fig:FSS}
\end{figure}

\mypar{Results.}
We report our FSS results in Fig.~\ref{fig:FSS} where we show the mIoU performance of all models. Our results show that pretraining on CrashCar101 yields significantly better results (+6.3-17.9\% mIoU at 1-shot and +4.4-7.0\% at 5-shot) compared to using a pre-trained model on COCO or ImageNet. As expected, when the amount of real data increases the performance gain decreases, nonetheless we still see a marginal performance gain for the 10-shot experiments. Both SegFormer and DeepLabv3 perform better when pre-trained on CrashCar101, which suggests that the improvement is independent of the model architecture. These results show that there is a smaller domain gap between CrashCar101 and CarDD than there is between COCO/ImageNet and CarDD. These results show the potential of the sim2real transfer of our dataset on the task of damage segmentation.

%% file: 6_Conclusion.tex
\section{Conclusion}

We proposed a procedural generation pipeline that creates damages on 3D cars.
We executed our pipeline and rendered the CrashCar101 synthetic dataset. 
We showed that without any special modification or any domain adaptation methods, our CrashCar101 dataset is useful for training a damage assessment system that performs damage segmentation and semantic part segmentation on real images. 
We hope that our work will enable more work in this direction and lead to a more powerful synthetic data generation pipeline able to deal with a variety of different incidents such as natural disasters and damage assessment models that can operate on various objects beyond vehicles.
%


\mypar{Acknowledgments.} D. Papadopoulos was supported by the DFF Sapere Aude Starting Grant ``ACHILLES''. We would like to thank Thanos Delatolas for proofreading.